\documentclass[a4paper,fleqn]{cas-sc}

\usepackage[authoryear]{natbib}

\usepackage{multirow}
\DeclareUnicodeCharacter{2061}{}

\def\tsc#1{\csdef{#1}{\textsc{\lowercase{#1}}\xspace}}
\tsc{WGM}
\tsc{QE}

\begin{document}
\let\WriteBookmarks\relax
\def\floatpagepagefraction{1}
\def\textpagefraction{.001}

\shorttitle{}    

\shortauthors{X. Wang et~al.}  

\title [mode = title]{Multimodal contrastive learning of urban space representations from POI data} 



\author[1]{Xinglei Wang}[
                        orcid=0000-0002-9824-7663
                        ]
\ead{xinglei.wang.21@ucl.ac.uk}

\credit{Conceptualization, Methodology, Software, Formal analysis, Data curation, Visualization, Writing - original draft, Writing - review \& editing}

\affiliation[1]{organization={SpaceTimeLab, Department of Civil, Environmental and Geomatic Engineering, University College London},
                city={London},
                postcode={WC1E 6BT}, 
                country={United Kingdom}}

\author[1]{Tao Cheng}[
orcid=0000-0002-5503-9813
]
\cormark[1]
\ead{tao.cheng@ucl.ac.uk}
\credit{Conceptualization, Resources, Supervision, Writing - review \& editing}

\author[2]{Stephen Law}[orcid=0000-0003-3184-572X]
\ead{stephen.law@ucl.ac.uk}
\credit{Conceptualization, Methodology, Supervision, Writing - review \& editing}

\author[1,4]{Zichao Zeng}[orcid=0009-0002-8975-875X]
\ead{zichao.zeng.21@ucl.ac.uk}
\credit{Conceptualization, Methodology, Writing - review \& editing}

\affiliation[2]{organization={Department of Geography, University College London},
                city={London},
                postcode={WC1E 6BT}, 
                country={United Kingdom}}

\author[3]{Lu Yin}[orcid=0000-0002-4075-0848]
\ead{l.yin@surrey.ac.uk}
\credit{Methodology, Supervision, Writing - review \& editing}

\author[1]{Junyuan Liu}[orcid=0009-0009-7194-6868]
\ead{junyuan.liu.22@ucl.ac.uk}
\credit{Formal analysis, Writing - review \& editing}

\affiliation[3]{organization={School of Computer Science and Electronic Engineering, University of Surrey},
                city={Surrey},
                postcode={GU2 7XH}, 
                country={United Kingdom}}

\affiliation[4]{organization={3DIMPact, Department of Civil, Environmental and Geomatic Engineering, University College London},
                city={London},
                postcode={WC1E 6BT}, 
                country={United Kingdom}}

\cortext[cor1]{Corresponding author}

\begin{abstract}
Existing methods for learning urban space representations from Point-of-Interest (POI) data face several limitations, including issues with geographical delineation, inadequate spatial information modelling, underutilisation of POI semantic attributes, and computational inefficiencies. To address these issues, we propose CaLLiPer (\underline{C}ontr\underline{a}stive \underline{L}anguage-\underline{L}ocat\underline{i}on \underline{P}r\underline{e}-t\underline{r}aining), a novel representation learning model that directly embeds continuous urban spaces into vector representations that can capture the spatial and semantic distribution of urban environment. This model leverages a multimodal contrastive learning objective, aligning location embeddings with textual POI descriptions, thereby bypassing the need for complex training corpus construction and negative sampling. We validate CaLLiPer’s effectiveness by applying it to learning urban space representations in London, UK, where it demonstrates 5 - 15\% improvement in predictive performance for land use classification and socioeconomic distribution mapping tasks compared to state-of-the-art methods. Visualisations of the learned representations further illustrate our model’s advantages in capturing spatial variations in urban semantics with high accuracy and fine resolution. Additionally, CaLLiPer achieves reduced training time, showcasing its efficiency and scalability. This work provides a promising pathway for scalable, semantically rich urban space representation learning that can support the development of geospatial foundation models. The implementation code is available at https://github.com/xlwang233/CaLLiPer.
\end{abstract}




\maketitle

\section{Introduction}

Accurate and comprehensive characterisation of urban spaces help us better understand how cities develop and function \citep{carmona2021public,gill2008characterising}, upon which we can further renovate and improve existing urban environment or future planning processes in order to tackle pressing challenges such as spatial inequalities \citep{nijman2020urban} and sustainability \citep{puchol2021urban}. Point of interest (POI) data have been widely used for characterising urban spaces, as they provide crucial information about both “where” (spatial) and “what” (platial) aspects of cities \citep{goodchild2020platial}, capturing fine-grained and up-to-date details depicting the land use composition, urban facilities distribution, and socioeconomic fabric of cities \citep{liu2020considering}. 

In today’s booming age of big data and artificial intelligence, urban studies increasingly rely on computational methods to model, simulate, and analyse urban environments. Consequently, there has been extensive research focused on deep representation learning of urban spaces from POI data. Such representation learning are performed on various geographical scales, ranging from locations \citep{hong2023context}, neighbourhoods \citep{huang2021learning,wang2020urban2vec}, to regions \citep{huang2022sppe,niu2021delineating} and city level \citep{huang2023hgi}. And the learned representations have found widespread applications in urban functional distribution mapping \citep{huang2022sppe,huang2023hgi}, land use classification \citep{jean2019tile2vec} socioeconomic indicator estimation \citep{jean2016combining}, footfall prediction \citep{feng2017poi2vec} and individual’s next location prediction \citep{hong2023context}. 

Existing urban space representation learning approaches are largely unsupervised or self-supervised, aiming at learning general purpose representations that can generalise to various downstream tasks. These methods propose to learn spatial co-occurrence patterns between POI categories and derive urban space representations as the aggregation of the POI category embeddings contained within \citep{huang2022sppe,huang2023hgi,yan2017itdl,yao2017sensing,zhai2019beyond}. These methods have several limitations:

First, the spatial locations of POIs have not been explicitly represented and combined with their semantic information in existing models, as no current learning algorithm effectively integrates both locational information (coordinates, "where") and semantic attributes (POI types, "what"). 

Second, these methods are subject to the geographical delineation of the urban space, which are often discretised areas like administrative regions or census tracts. These delineations are either predetermined during the pre-training stage or specified later in the aggregation stage. It is challenging to determine the ideal spatial resolution \citep{niu2021delineating} as a finer one might result in no POIs within specific areas, while a coarser resolution might introduce ecological fallacy issue \citep{robinson2009ecological}. Moreover, it is hard for representations learned on one spatial scale to be applied to another scale and it would often require retraining the model.

Third, existing methods fail to fully utilise the rich textual information of POI data. They learn the semantic representations of POIs via pure spatial co-occurrence patterns, based on the hypothesis that categories occurring with similar spatial contexts tend to have similar semantic properties \citep{bing2022pre}. However, these implicit learning approaches neglect the fact that POI labels, being natural language descriptions per se, carry inherent semantic meanings that can be modelled from a computational linguistics perspective.

Lastly, existing state-of-the-art methods are not efficient in terms of training. Most of them are based on word embedding approaches \citep{mikolov2013efficient,mikolov2013distributed} used in natural language processing (NLP) domain. The construction of POIs co-occurrences corpus requires sophisticated design of spatial context . Moreover, some methods demand complex negative sampling strategies to facilitate contrastive learning \citep{huang2022sppe,huang2023hgi}. The high time complexities (e.g., $O(n^2)$) of these methods incur prolonged training time.

To address these challenges, this paper proposes a novel method for learning urban space representations, which we call \textbf{CaLLiPer}, for \textit{\underline{C}ontr\underline{a}stive \underline{L}anguage-\underline{L}ocat\underline{i}on \underline{P}r\underline{e}-t\underline{r}aining}. The core idea is to explicitly embed both locational and semantic information extracted from POI data into effective urban space representations. We propose a location encoder that embeds locations (coordinates) into vector representations that can integrate with the vector representations of the semantics of POIs. Such an ability is achieved through a location encoding process guided by a multimodal contrastive learning objective. Unlike other methods that embed discrete geospatial objects such as POIs or regions, our model treats the entire urban environment as a continuous space, enabling inductive learning that does not require retraining to extrapolate to unseen geospatial entities \citep{mai2022review}. Additionally, we use pre-trained text encoders to embed POI descriptive texts into textual representations, which are matched with the location embeddings of the corresponding POI coordinates, aligning the location embedding space with the well-trained language embedding space. Moreover, the construction of training data, i.e., (coordinate, text) pairs, is straightforward and much simpler than building a co-occurrence corpus. The multimodal contrastive learning objective also eliminates the need for extra negative sampling procedures, making the model highly efficient. 

To verify the effectiveness of our method, we apply CaLLiPer to learning urban space representations in London, UK, which are then used in two downstream tasks: urban land use classification and socioeconomic status distribution mapping. We compare our method with state-of-the-art models in terms of predictive performance on downstream tasks and training efficiency. Experimental results show that CaLLiPer can achieve 5\% - 15\% performance improvement on downstream tasks compared to the best-performing baselines, while having relatively short training time. Additionally, we conduct qualitative analysis of the representations to gain more understandings of our model.

We briefly summarise our contributions as follows: 

\begin{itemize}
    \item We advance urban space representation learning by explicitly integrating locational and semantic information from POI data into comprehensive, scalable representations, enabling the joint modelling of spatial (coordinates) and platial (texual labels) information. To the best of our knowledge, this is the first study that combines these two unique modalities through multimodal contrastive learning.
    \item Our approach introduces several methodological innovations: treating cities as continuous spaces via location encoding, enabling flexible, scale-free applications without retraining; leveraging pre-trained NLP text encoders to capture POI semantics more effectively; and using multimodal contrastive learning to align spatial and semantic representations. This approach not only enhances spatial-semantic learning but also reduces the need for complex corpus design and negative sampling, improving computational efficiency.
    \item Extensive experiments show that our proposed model outperforms state-of-the-art models by 5 - 15\% in predictive accuracy across two downstream tasks while reducing training times, proving its effectiveness in real-world urban analysis.
\end{itemize}

Following this introduction, we review related work in Section \ref{sec:lit_review}. In Section \ref{sec:notation}, we introduce notations used in the paper the formulation of POI-based urban representation learning. Section \ref{sec:method} elaborates on the detailed architecture and mathematical formulation of the proposed CaLLiPer model, as well as the deployment of the learned representations in downstream tasks. We present the experimental setup in Section \ref{sec:exp}, followed by results and analysis in Section \ref{sec:res}. We summarise the major findings and discuss the potentials of CaLLiPer for geospatial foundation models in Section \ref{sec:conclustion_discussion}.

\section{Related work}
\label{sec:lit_review}

Our work builds on best practices from a range of studies. We adopt location encoding in our urban space encoder and utilise well-trained text encoders from the NLP domain to extract rich semantic information from textual descriptions of POIs. Additionally, we employ a contrastive multimodal representation learning process to encode this semantic information into the location encoder. In the following subsections, we review not only existing methods for urban space representation learning but also the relevant techniques incorporated into our model.

\subsection{Learning urban space representations with POIs}

There has been extensive research on learning numerical representations of urban spaces. Early studies utilised topic models to derive these representations, with TF-IDF \citep{sparck1972tfidf} and LDA \citep{blei2003latent} being the most commonly applied. These models have been used for region function discovery \citep{yuan_discovering_2012} and semantic enrichment of places \citep{cheng2018grouping,shen2018profiling}. Later, following the success of Word2Vec models in modelling the distributed representation of words \citep{mikolov2013distributed}, numerous studies adapted this approach to learn distributed representations of POIs and aggregate the POI embeddings to various urban spaces. The general workflow of these methods consists of two steps: first, learning POI category embeddings, and second, deriving the urban space representations by aggregating the POI category embeddings within specific areas. 

Learning POI category embeddings requires a POI co-occurrence corpus that can constructed in various ways. \citet{yao2017sensing} utilised shortest path algorithm to traverse all POIs in a given area and extract the co-occurrence corpus from the resulting POI sequences. \citet{yan2017itdl} proposed the Place2Vec model, which identifies context POIs using a K-nearest-neighbor (KNN) method. \citet{niu2021delineating} constructed the corpus along road networks while \citet{huang2022sppe} built a POI network through Delaunay triangulation and captured the spatial co-occurrence patterns through random walks on this network. 

For the aggregation of POI embeddings into urban space representations, most studies have employed an average pooling function. \citet{niu2021delineating} used the Doc2Vec model to obtain urban region embeddings as a by-product of the POI embeddings. \citet{huang2022sppe} used a long short-term memory (LSTM) model and attention mechanisms to aggregate POI embeddings to urban region embedding through a supervised training objective. More recently, \citet{huang2023hgi} employed multi-head attention and graph neural networks (GNNs) in their aggregation process and introduced hard negative sampling strategy and a contrastive training objective to facilitate fully unsupervised learning. 

We argue that the POI category embedding aggregation-based paradigm for learning urban space representations have several limitations. First, it emphasises learning the representation of POI category rather than directly addressing the primary challenge of capturing spatial variations in POI distribution, which is central to urban space representation learning. Additionally, these methods focus on learning POI semantic embedding via modelling the spatial context of different POI categories, failing to fully utilise the textual description of POIs. Furthermore, the second step of the learning paradigm – aggregating POI embeddings within specific areas – relies on the geographical delineation of urban spaces. The problem is that deciding the appropriate spatial resolution for these areas is challenging, and slight changes in delineation can significantly alter the resulting representations. 

In terms of computational efficiency, probabilistic topic models (e.g., TF-IDF and LDA) are highly efficient as they do not require training. However, deep representation learning approaches involve constructing co-occurrence corpora and executing sophisticated training processes, both of which increase computation time. 

Addressing these limitations, our method directly learns the spatial-semantic representation of urban spaces by integrating location encoding with semantic encoding from POIs through contrastive learning, thereby improving both modelling accuracy and efficiency. 

\subsection{Location encoding}\label{sec:lit_loc_encoding}

Location encoding refers to the process of encoding point locations into an embedding space so that these location embeddings can be readily used in downstream neural network modules \citep{mai2022review}, with various aims like geographic prior modelling \citep{chu2019geo,klemmer2023satclip,mac2019presence} or spatial context modelling \citep{mai2020iclr}, etc.

The motivation of such an idea was first comprehensively articulated in \citep{mai2022review}, in which the authors provided a general conceptual framework that unifies the formulation of location encoding methods. Following their formulation, location encoding methods generally take the form of $y=\text{NN}(\text{PE}(\lambda,\phi))$, where a geographical or projected coordinate $(\lambda,\phi)$ is processed through a parametric positional encoding (PE) and a neural network (NN). 

Depending on specific PE functions, location encoding can be performed on either planar or spherical geometries at different spatial scales. Over recent years, a series of PE methods have been proposed, including Wrap \citep{mac2019presence}, Grid and Theory \citep{mai2020iclr}, Sphere* \citep{mai2023sphere2vec}, and Spherical Harmonics (SH) \citep{russwurm2024sh}, etc.

Theoretically, location encoding can embed continuous urban spaces into an embedding space, enabling representations at extremely fine scales (e.g., metre or even sub-metre level). Therefore, it has great potential for addressing the ecological fallacy that can arise under coarse spatial scales. Learning a continuous encoding function instead of a discrete embedding matrix also makes the model highly inductive and generalisable. 

Practice of location encoding has been found in a variety of downstream applications, including geo-aware image classification \citep{mac2019presence}, POI classification \citep{mai2020iclr}, geographic question answering \citep{mai2020se}, etc. However, it has not yet been explored for learning urban space representations with POIs in a multimodal setting, which is the one of the research gaps we are addressing in this paper. 

\subsection{Text embedding}

Text embedding models encode the semantic content of natural language into vector representations, facilitating various natural language processing (NLP) tasks such as semantic textual similarity, information retrieval, question answering, and clustering \citep{behnamghader2024llmvec}. Early approaches often employed weighted averages of pre-trained word embeddings to measure semantic similarity \citep{arora2017simple,pennington2014glove}. With the development of pretrained language models, like BERT \citep{devlin2019bert}, more advanced methods, such as Sentence-BERT \citep{reimers2019sbert} and SimCSE \citep{gao2021simcse}, have been developed to fine-tune BERT on natural language inference datasets. These models use encoder-only or encoder-decoder architectures, while decoder-only language models have also shown strong performance in producing high-quality text embeddings \citep{muennighoff2022sgpt}. Recently, the rise of large language models (LLMs) \citep{touvron2023llama,touvron2023llama2,zhao2023survey} has encouraged researchers to apply them to text embedding tasks \citep{behnamghader2024llmvec,ma2024fine,wang2023improving}, achieving new state-of-the-art performance on benchmark datasets.

Our proposed CaLLiPer model leverages text embedding models to embed the textual descriptions of POIs into high-quality vector representations, taking advantages of their advanced capacity to capture the semantics of text. 

\subsection{Contrastive learning and multimodal learning}
\label{sec:2.4}

Contrastive learning is a self-supervised learning technique used to learn representations by contrasting positive and negative pairs of examples. It aims to bring similar (positive) examples closer in the representation space and push dissimilar (negative) examples farther apart \citep{chen2020simple}. This technique has been widely applied in various research domains, including computer vision \citep{chen2020simple,he2020momentum}, NLP \citep{gao2021simcse,gunel2021supervised}, recommendation systems \citep{chen2022intent} and more. This technique has also been applied in representation learning of urban spaces. In a pioneering study on POI-based urban region representation learning, \citet{liu2018efficient} treated each region as an “image” where pixels are filled with POIs. For each anchor region, positive samples were generated by random removal, addition, and shifting of POIs, while negative samples are non-overlapping regions or augmentations with larger perturbations (hard negative samples). More recently, \citet{huang2023hgi} proposed generating city-region and region-POI negative sample pairs to learn POI embeddings and region embeddings through maximising the mutual information among the POI-region-city hierarchy. These single modality learning methods require dedicated positive and negative samples to enable contrastive learning within the same modality. 

The development of jointly learning representations for multiple modalities have opened new avenue for contrastive learning. One pioneering and groundbreaking model for multimodal representation learning is Contrastive Language-Image Pre-training (CLIP) model \citep{radford2021clip}. CLIP is trained on large datasets of images and their associated textual descriptions (e.g., captions). Each image is naturally linked to a positive text description (the actual caption) and many negative examples (captions for other images) so that no extra negative sampling process is required. CLIP demonstrated impressive zero-shot image classification ability and have inspired subsequent research into representation learning of urban environment. \citet{yan2024urbanclip} proposed the UrbanCLIP model to pre-train an image encoder for urban regions. They generated textual descriptions for satellite images from well-trained LLMs and deployed a CLIP-like process to train the image encoder. However, unlike other pre-trained urban space embeddings that can be used off-the-shelf, this approach requires satellite images as input to obtain region representations.   

Several most recent works have treated location as a unique modality and incorporated it into multimodal contrastive learning. GeoCLIP \citep{vivanco2024geoclip} was developed for the task of image-based geo-localisation, which is essentially an image-to-coordinate retrieval problem.  \citet{klemmer2023satclip} proposed SatCLIP, which aims to distil environmental ground conditions from satellite images into a global location encoder that operates across the whole planet. This pre-trained location encoder only requires geographic coordinates to generate location embeddings. While we draw from their approach of integrating different modalities, we diverge by using text from POI data to extract semantic information for our location encoder. The motivation is twofold. First, satellite images often lack detailed semantics and may not always be distinguishable; for example, two urban areas with similar visual textures may have contrasting functions. Using POI data allows us to capture the richer and more nuanced people-centric information compared to the general observations from aerial photographs. Second, while satellite images are suitable for constructing location-image pairs on a global scale, they are not ideal for constructing such kind of training data for smaller, local areas with fine resolution. This is because it is unlikely for two nearby locations, just metres apart, to have distinct, non-overlapping satellite images. 

\section{Notations and Problem Statement}
\label{sec:notation}

A set of POIs is denoted as $\mathcal{P}=\{p_1,p_2,...,p_i,...,p_M\}$, containing a total of $M$ POIs. Each POI $p_i=(x_i^L,x_i^S )$ includes location $x_i^L$ and semantic properties $x_i^S$. The location of a POI is typically a pair of coordinates $x_i^L=(\lambda_i,\phi_i)$ while the semantic properties $x_i^S$ are generally labels that indicate the POI’s functions or attributes. For example, a POI could be represented as ((536366.19, 190390.01), Retail). Existing approaches treat the POI semantic labels as purely categorical values, i.e., $x_i^S \equiv x_i^C$. However, in our proposed model, these labels are treated as textual descriptions, i.e., $x_i^S \equiv x_i^T$.

A certain urban space is denoted as $u$. Depending on specific scales, urban spaces can range from individual buildings, neighbourhoods to districts and regions. The goal of urban space representation learning is to develop a model F that can embed an arbitrary urban space u into a meaningful representation $e_u=F(u)$, which encodes the rich semantic features of the local urban environment.

For methods based on POI category embeddings, the learning target is an embedding matrix of the POI categories, denoted as $E$. And the representation of a space $u$ is then expressed as an aggregation of the POIs therein:

\begin{equation}
    e_u=Agg_{p_i \in P_u} (E(x_i^C))
\end{equation} where $\mathcal{P}_u$ is the subset of POIs located within the space $u$, $E$ is the embedding matrix of POI categories, and $Agg$ denotes the aggregation function, which can vary depending on the method.

In contrast, our proposed model directly embeds the entire urban environment using a location encoder $f^L$. And thus, the learned representation of a space $u$ can be written as: 
\begin{equation}
    e_u=f^L (g(u))
\end{equation} where $g(u)$ denotes the point location of the space $u$, which normally is the centroid.

\section{Method}
\label{sec:method}

\subsection{Model overview}
\label{sec:method_overview}

\begin{figure}
    \centering
    \includegraphics[width=0.98\linewidth]{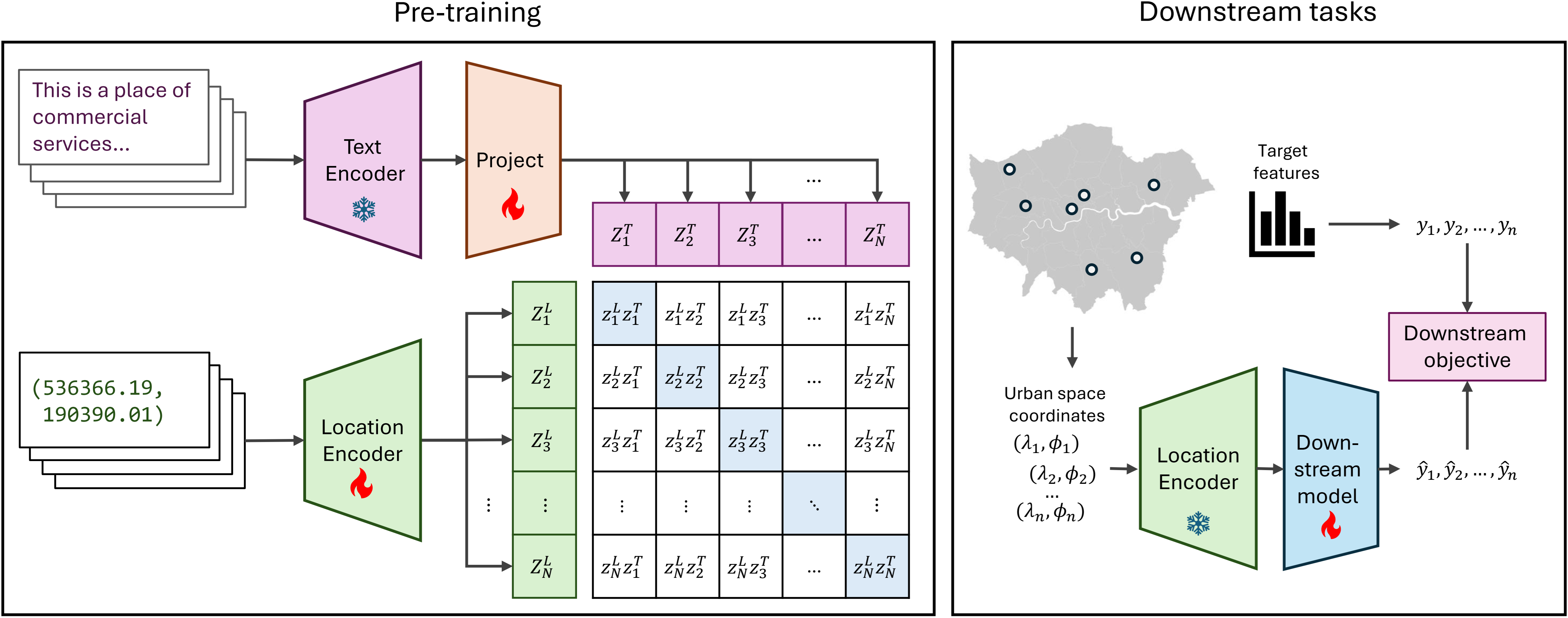}
    \caption{The CaLLiPer pre-training and application pipeline.}
    \label{fig:model_overview}
\end{figure}

Figure \ref{fig:model_overview} shows the overall framework of the proposed CaLLiPer model (left pane) and the workflow of downstream applications (right pane). The working mechanism of our method comprises two stages: pre-training of the CaLLiPer model and applications in downstream tasks.

The CaLLiPer model consists of three main components: (1) a location encoder $f^L$ that embeds individual coordinates into higher dimensional space; (2) a text encoder $f^T$ for extracting semantic features from POI’s natural language descriptions; and (3) a projection layer $f^P$ that projects the output of text encoder to match the dimensions of the location embeddings. The details of each model component are introduced in Section \ref{sec:method_loc_encoder} to Section \ref{sec:method_proj_layer}. 

In the pre-training stage, the location encoder embeds a batch of $N$ POI coordinates $\boldsymbol{x}^L \in \mathbb{R}^{N \times 2}$ into location embeddings $\boldsymbol{z}^L \in \mathbb{R}^{N \times d}$ in a $d$-dimensional space. Simultaneously, the corresponding POI textual descriptions $x^T \in \mathbb{R}^{N \times t}$ ($N$ sentences with arbitrary length $t$) pass through the text encoder and projection layer and be mapped to text embeddings $z^T \in \mathbb{R}^{N \times d}$. The mathematical formulations are as follows:

\begin{equation}
    f_{\Theta ^L}^L (x^L ) = z^L \in \mathbb{R}^{N \times d}
\end{equation}

\begin{equation}
    f_{\Theta ^P}^P (f^T (x^T))= z^T \in \mathbb{R}^{N \times d}  
\end{equation}

The parameters of the location encoder and the projection layer, i.e., $\Theta^L$ and $\Theta^P$, are optimised, while the text encoder is frozen and do not receive parameter update. We utilise the simple yet highly effective bi-directional InfoNCE as the training objective \citep{radford2021clip}:

\begin{equation}
    \mathcal{L}(\Theta^L,\Theta^P)=-\frac{1}{2N}[\sum_{i=1}^{N}\log\frac{\exp⁡(z_i^L z_i^T/\tau)}{\sum_{j=1}^{N}\exp⁡(z_i^L z_j^T / \tau)} + \sum_{i=1}^{N}\log\frac{\exp⁡(z_i^T z_i^L/ \tau)}{\sum_{j=1}^{N}\exp⁡(z_i^T z_j^L / \tau)}]
\end{equation} where $\tau$ is a temperature hyperparameter. 

Once the pre-training is finished, the location encoder can be used in various downstream tasks in a “plug and play” manner, producing the embeddings for any urban spaces without further training, while an additional lightweight downstream model $f^D$ is optimised to obtain the final target features. 

\begin{equation}
\label{eq:downstream}
    \hat{y}_{i^D}=f_{\Theta^D}^D (f_{\Theta^L}^L (x_{i^D}^L))
\end{equation} where $\Theta^D$ is the trainable parameters of the downstream model, $i^D$ indicates the $i$-th sample in downstream task $D$, and $\hat{y}_{i^D}$ is the corresponding prediction result.

The details of each model component are introduced in the following subsections. 

\subsection{Location encoder}
\label{sec:method_loc_encoder}

As introduced in Section \ref{sec:lit_loc_encoding}, all location encoding approaches take the form of 

\begin{equation}
    \text{NN}(\text{PE}(\lambda,\phi))
\end{equation} where a parametric positional encoding (PE) projects the raw coordinates $(\lambda,\phi)$ into a higher-dimensional encoding space, followed by a neural network (NN) that further encodes the target feature distributions. 

\begin{figure}
    \centering
    \includegraphics[width=0.4\linewidth]{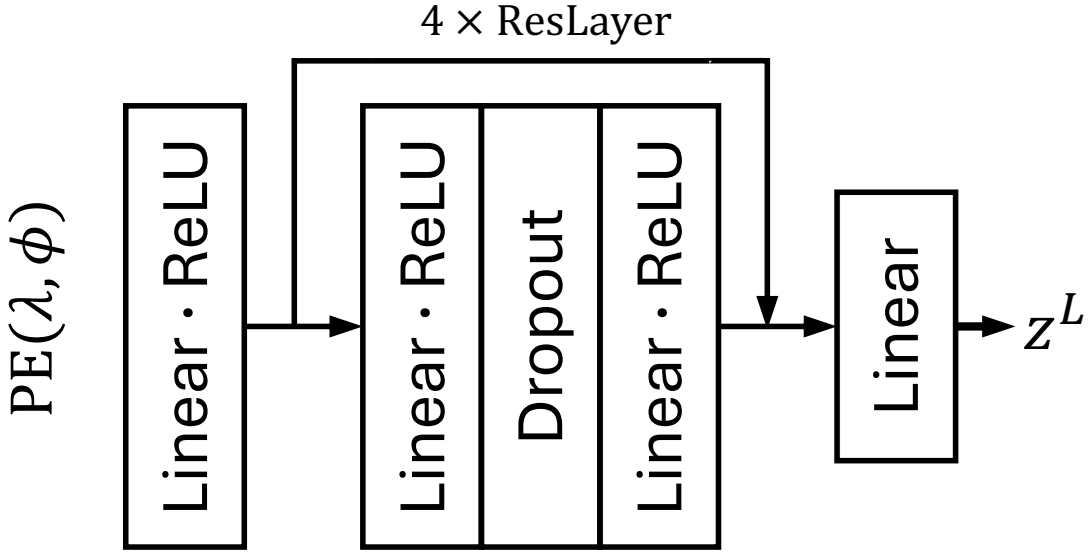}
    \caption{The architecture of FC-Net.}
    \label{fig:fc_net}
\end{figure}

In this work, we employ Grid \citep{mai2020iclr} as the PE, which is motivated by several considerations. First, our goal is to embed city-wide coordinates into a high dimensional space, and thus, we use projected coordinates on 2D planes rather than geographical coordinates that work on a spherical surface. Secondly, we aim for the location encoder to capture multi-scale spatial information while remaining computationally efficient. Grid operates effectively on projected coordinates, resolves multiple scales, and is more efficient than alternative methods like Theory \citep{mai2020iclr} without sacrificing the performance. Therefore, we choose it as the PE method. As for the design of NN, we adopt a fully connected residual network named FC-Net (Figure \ref{fig:fc_net}), as it is a performant and widely used neural architecture in location encoding \citep{russwurm2024sh}.   

The mathematical formulation of Grid \citep{mai2020iclr} is as follows:

\begin{equation}
    \text{PE}(\lambda,\phi)=\bigcup_{s=0}^{S-1}(\cos \frac{\lambda}{\alpha_s} ,\sin \frac{\lambda}{\alpha_{s}},\cos \frac{\phi}{\alpha_s},\sin \frac{\phi}{\alpha_s})
\end{equation}

\begin{equation}
    \alpha_{s}=r_{\text{min}} \cdot (\frac{r_{\text{max}}}{r_{\text{min}}})^{\frac{1}{s-1}}
\end{equation} where $r_{\text{min}}$ and $r_{\text{max}}$ are the minimum and maximum radii, respectively, and $S$ is the number of scales. These hyperparameters control the resolutions of the muti-scale encoding of the coordinates. 

\subsection{Text encoder}
\label{sec:method_text_encoder}

In order to obtain expressive textual representations that effectively encode the semantic features of POIs, we propose using off-the-shelf language models that have been pre-trained on massive amounts of high-quality textual data as the text encoder in CaLLiPer. Specifically, we utilise Sentence-BERT \citep{reimers2019sbert} (also known as SBERT or Sentence Transformers) and Llama3-8B \citep{dubey2024llama3}, which are encoder-only and decoder-only Transformer architectures, respectively. It is worth noting that Llama3-8B, one of the most recent and performant LLMs, is significantly larger in size compared to Sentence-BERT. 

Inspired by \citep{radford2021clip}, we design prompt templates to further enhance the text encoding process to better capture the semantics of POIs. Generally, we use “a place of \{label\}” as a default prompt template, which is highly customisable, depending on the specific labels of the POIs. For example, some POIs of the “retail” type may also include specific goods information, and in such cases, we use “a place that sells \{goods\_type\}”. Additionally, some POI data might include a hierarchy of labels and the prompt template can be adjusted accordingly, such as “a place of \{label\_1\}, a type of \{label\_2\}”. More details regarding prompt templates are described in Appendix \ref{sec:appendix_prompt_template}.

\subsection{Projection layer}
\label{sec:method_proj_layer}

The projection layer is essentially a linear layer that projects the text embeddings (output of the text encoder) to the same dimension as the location embeddings, facilitating the construction of the contrastive learning objective between these two modalities. This kind of projection layer is commonly used for jointly embedding multiple modalities in contrastive pre-training \citep{radford2021clip} and for aligning two modalities in fine-tuning processes \citep{jin2024time,liu2024llava}.

\subsection{Applying to downstream tasks}

After the pre-training of CaLLiPer is completed, the pre-trained location encoder can be applied to various downstream tasks that are influenced by the spatial-semantic distribution information. As shown in the right pane of Figure \ref{fig:model_overview}, the representations of arbitrary urban spaces in downstream tasks are generated by inputting their point locations into the pre-trained location encoder. These representations then serve as the input (or part of the input) to a downstream model $f^D$, which undergoes a supervised leaning process to predict target variables. Note that the pre-trained location encoder is frozen in this stage, with only the downstream model being trained. This process is formulated in Equation \ref{eq:downstream}. 

\section{Experimental setup}
\label{sec:exp}

We choose London as the study area to evaluate the proposed model. London is the capital and largest city in the UK, and an important international city in the world. It not only has a diverse urban spatial configuration, but also a diverse sociodemographic population, making it an ideal testbed to conduct our experiments. 

We evaluate the learned urban space representations using two downstream tasks: land use classification (LUC) and socioeconomic status distribution mapping (SDM). LUC is widely recognised as a standard task for verifying urban representation learning, especially in capturing urban functional distributions \citep{huang2022sppe,huang2023hgi,zhai2019beyond}. SDM, on the other hand, is particularly relevant because socioeconomic status (SES) and the urban environment are strongly intertwined, with SES affecting access to essential urban resources such as education and healthcare \citep{mcmaughan2020socioeconomic}. In turn, the distribution of these resources shapes living conditions and reinforces inequality and segregation \citep{useche2024spatial}.

We introduce the data used in the downstream tasks testing, comparison methods, implementation details in the following subsections. 

\subsection{Data}

\textbf{POI data}. The POI data, comprising 339,956 POIs in the Greater London Area, was obtained from Ordnance Survey through Digimap\footnote{https://digimap.edina.ac.uk/} using an educational licence. The dataset is maintained and updated regularly; we used the March 2022 version in the experiment. The POI classification scheme follows a three-tier hierarchy with 9 groups, 52 categories and 616 classes\footnote{The detailed classification scheme is available at https://www.ordnancesurvey.co.uk/documents/product-support/user-guide/points-of-interest-classification-schemes-v3.3.pdf}. 

\textbf{Land use dataset}. The land use dataset was obtained from EDINA Verisk Digimap Service\footnote{https://digimap.edina.ac.uk/roam/map/verisk}, with high-level classification proposed in the National Land-Use Database\footnote{https://www.gov.uk/government/statistics/national-land-use-database-land-use-and-land-cover-classification}. To construct a dataset that is suitable for the LUC task, we sampled a number of points with 200-metre radius buffer and further adjusted the number of samples for certain land use types to account for the data imbalance. The resulting dataset contains 6697 samples, with the counts of different land use types shown in Table \ref{tab:luc_data}. 

\begin{table}[]
\caption{The basic statistics of the land use dataset used in the LUC task.}
\begin{tabular}{ll}
\toprule
Land use type & Count \\
\midrule
High density residential with retail and commercial   sites & 895 \\
Low density residential with amenities (suburbs and   small villages / hamlets) & 895 \\
Medium density residential with high streets and   amenities & 895 \\
Agriculture - mainly crops & 895 \\
Deciduous woodland & 895 \\
Recreational land & 847 \\
Principle transport & 603 \\
Industrial areas & 270 \\
Large complex buildings various use   (travel/recreation/ retail) & 254 \\
Water & 248 \\
\bottomrule
\end{tabular}
\label{tab:luc_data}
\end{table}

\textbf{National Statistics Socio-economic Classification (NS-SeC) dataset}. The NS-SeC dataset categorises a person's socio-economic position based on their occupations and other job characteristics (see Table \ref{tab:ns_sec_data} for the detailed classification scheme). This dataset is part of the 2021 Census and was acquired from the Office for National Statistics (ONS)\footnote{https://www.ons.gov.uk/}. We use data organised by Lower-layer Super Output Areas (LSOAs), which are a kind of lower-level geographical units for census statistics, each comprising 400 to 1200 households and a resident population of 1000 to 3000 people. The NS-SeC dataset used in our experiment consists of 4994 samples, corresponding to the 4994 LSOAs in London. 

\begin{table}[]
\caption{The detailed classification scheme of NS-SeC from ONS.}
\begin{tabular}{l}
\toprule
National Statistics Socio-economic Classification   (NS-SeC) (10 categories) \\
\midrule
L1, L2 and L3: Higher managerial, administrative and   professional occupations \\
L4, L5 and L6: Lower managerial, administrative and   professional occupations \\
L7: Intermediate occupations \\
L8 and L9: Small employers and own account workers \\
L10 and L11: Lower supervisory and technical   occupations \\
L12: Semi-routine occupations \\
L13: Routine occupations \\
L14.1 and L14.2: Never worked and long-term   unemployed \\
L15: Full-time students \\
\bottomrule
\end{tabular}
\label{tab:ns_sec_data}
\end{table}


\subsection{Comparison methods}

We compare our model with a range of unsupervised learning models, including a baseline method (Random); topic modelling-based approaches (TF-IDF and LDA); approaches that first learn POI category embeddings and then aggregate them within areas (Place2Vec, Doc2Vec, SPPE, HGI); and a location encoding-based method (Space2Vec). These methods are introduced as follows:

\begin{itemize}
    \item Random: This method randomly initialises embeddings for urban spaces, serving as the lower bound performance.
    \item TF-IDF \citep{sparck1972tfidf}: A statistical measure used in text mining to evaluate the importance of a word in a document relative to a collection of documents. It has been adopted in many studies to characterise the semantics of urban spaces \citep{niu2021delineating,shen2018profiling,zhai2019beyond}. 
    \item LDA \citep{blei2003latent}: A generative probabilistic model used in topic modelling that identifies underlying topics in a corpus, which has also been used in many studies.
    \item Place2vec \citep{yan2017itdl}: This method incorporates spatial co-occurrence information using a KNN sampling strategy and distance decay to learn POI category embeddings, subsequently creating region embeddings by averaging the POI category embeddings within each region.
    \item Doc2Vec \citep{niu2021delineating}: This approach constructs POI category co-occurrences using a KNN strategy, treating each POI as a “word” and each urban region as a “document” to jointly train embeddings for both POI categories and regions. The aggregation of POIs into regions occurs simultaneously during the model training process.
    \item SPPE (Semantics preserved POI embedding) \citep{huang2022sppe}: This method integrates both spatial co-occurrence information and categorical semantics, creating POI category embeddings using random walks combined with a manifold learning algorithm. In aggregating POI category embeddings to region embeddings, the authors proposed either a supervised aggregation function coupling LSTM and attention mechanisms or an unsupervised average pooling. For fair comparison with other unsupervised learning methods, mean pooling was used to generate region embeddings. 
    \item HGI \citep{huang2023hgi}: This method extends the POI category embedding approach of SPPE \citep{huang2022sppe} by incorporating a GNN aggregation process to enhance the representations at both region and city levels. The model is optimised by maximising mutual information across the POI-region-city hierarchy. 
    \item Space2Vec \citep{mai2020iclr}: A location encoding method trained using an encoder-decoder architecture on POI data. The encoder embeds POI coordinates into location embeddings, which are then passed to the decoder to reconstruct POI features (categories). The unsupervised training maximises the log likelihood of observing the true point at its position among all points, with negative sampling employed to improve training efficiency.
\end{itemize}
 
Other representation learning methods that rely on data sources such as human mobility data, street-view images, or satellite images are excluded from this study, as they are unsuitable for comparison with models based solely on POI data. However, we provide the performance of a model pre-trained on satellite images in Appendix \ref{sec:appendix_satclip}, serving as a supplementary comparison.

\subsection{Implementation details}

\subsubsection{Computing environment}
All the experiments were conducted on a server equipped with an AMD Ryzen Threadripper PRO 5975WX and one Nvidia RTX A6000 GPU, running Ubuntu 22.04.4 LTS. 

\subsubsection{Pre-training models}
To ensure a fair comparison, we set the dimension for urban space representation as 128 across all methods. For CaLLiPer, we tuned the hyperparameters via grid search, resulting in the following final setttings: $\lambda_{\text{min}}=100$, $\lambda_\text{max}=10000$, $S=32$, and a hidden dimension of 256 for FC-Net. We implemented two versions of CaLLiPer model, each using a different text encoder: CaLLiPer-SenTrans, which employs Sentence Transformer, and CaLLiPer-Llama, which adopts Llama3-8B as its text encoder.

TF-IDF, LDA and Doc2Vec were implemented using the Gensim\footnote{https://radimrehurek.com/gensim/} library. All the deep learning-based models were implemented using PyTorch 2.2.2 \citep{paszke2019pytorch}. The corresponding open-source code for SPPE\footnote{https://github.com/RightBank/Semantics-preserved-POI-embedding}, HGI\footnote{https://github.com/RightBank/HGI}, and Space2Vec\footnote{https://github.com/gengchenmai/space2vec} were used in the implementation. For methods that require pre-defined spatial division of the urban space (i.e., Doc2Vec and HGI), we used LSOAs as the division. For Place2Vec and SPPE, which aggregate POI category embeddings, we used mean pooling as the aggregation approach. More implementation details regarding SPPE, HGI and Space2Vec can be found in Appendix \ref{sec:appendix_imp_sppe_hgi} and \ref{sec:appendix_imp_space2vec}.

\subsubsection{Downstream training and evaluation metrics}
Two downstream models were used to evaluate the pre-trained urban space representations. The first is a linear model, essentially a one-layer neural network for linear probing of the pre-trained representations. The second is a non-linear model, specifically a multi-layer perceptron (MLP) with a single hidden layer. 

For the LUC task, which is a multi-class classification task, we used cross-entropy loss to train the downstream models. The evaluation metrics are standard for classification tasks: Precision, Recall, and F1 score, macro-averaged over classes. For these metrics, higher values indicate better performance.

For the SDM task, a regression task, mean squared error loss was applied for the downstream training. For evaluation metrics, we follow prior studies \citep{huang2022sppe,huang2023hgi} and treat this task as a label distribution learning problem \citep{geng2016label}, selecting three representative measures: (1) L1 distance:$\sum_{k=1}^{m}|\hat{y}_i^{s_k}-y_i^{s_k}|$; (2) Chebyshev distance: $ max_k|\hat{y}_i^{s_k}-y_i^{s_k}|$
(3) KL divergence: $\sum_{k=1}^{m}y_i^{s_k}\log⁡(y_i^{s_k}/\hat{y}_i^{s_k})$, where $\hat{y}_i^{s_k}$ is the estimated proportion of the socioeconomic status $s_k$ that the $i$-th LSOA bears, and $y_i^{s_k}$ is the corresponding ground truth. For these metrics, lower values indicate better performance.

For each task, we split the dataset into training, validation, and testing sets in a 7:1:2 ratio. We trained the downstream model five times, each with a different random seed, and reported the mean and standard deviation across these five runs. Additionally, we monitored performance on the validation set and stopped training if the performance did not improve over 10 epochs. 

\section{Results and analyses}
\label{sec:res}

\subsection{Performance on downstream tasks}

\subsubsection{Land use classification}
\label{sec:luc_res}

\begin{table}[]
\caption{Performance comparison on the LUC task. The best and second-best performance are marked in \textbf{bold} and \underline{underlined}, respectively. For better readability, all metrics are scaled by a factor of $10^{-2}$.}
\label{tab:luc_res}
\begin{tabular}{l|ccc|ccc}
\toprule
\multirow{2}{*}{Model} & 
\multicolumn{3}{c|}{Linear} & \multicolumn{3}{c}{MLP} \\
 & Precision ↑ & Recall ↑ & F1 Score ↑ & Precision ↑ & Recall ↑ & F1 Score ↑ \\
\midrule
Random & 9.6 ± 0.7 & 10.3 ± 0.5 & 9.7 ± 0.5 & 8.8 ± 1.3 & 10.3 ± 0.3 & 9.0 ± 0.3 \\
TF-IDF & 31.5± 0.4 & 32.2 ± 0.2 & 31.3 ± 0.3 & 31.8 ± 0.6 & 33.3 ± 0.5 & 31.7 ± 0.6 \\
LDA & 30.8 ± 0.3 & 29.1 ± 0.2 & 28.4 ± 0.2 & 31.5 ± 1.1 & 30.4 ± 0.7 & 29.2 ± 0.9 \\
Place2Vec & 30.9 ± 0.8 & 26.1 ± 0.7 & 26.3 ± 0.7 & 35.1 ± 1.2 & 32.7 ± 1.0 & 32.4 ± 1.2 \\
Doc2Vec & 32.4 ± 0.4 & 28.2 ± 0.1 & 28.0 ± 0.1 & 34.9 ± 0.9 & 33.8 ± 0.5 & 32.7 ± 0.6 \\
SPPE & 30.5 ± 0.4 & 27.0 ± 0.2 & 26.6 ± 0.2 & 34.5 ± 0.9 & 32.9 ± 0.7 & 32.2 ± 0.5 \\
HGI & 33.0 ± 0.5 & 30.0 ± 0.6 & 29.9 ± 0.6 & 33.6 ± 0.5 & 32.0 ± 0.9 & 31.6 ± 0.7 \\
Space2Vec & 28.6 ± 0.6 & 28.5 ± 0.8 & 27.4 ± 0.7 & 29.6 ± 0.6 & 28.9 ± 0.52 & 27.8 ± 0.3 \\
\midrule 
CaLLiPer-SenTrans & $\mathbf{37.2 \pm 0.4}$ & $\mathbf{36.5 \pm 0.6}$ & $\mathbf{36.0 \pm 0.3}$ & $\underline{38.4 \pm 1.5}$ & $\mathbf{36.8 \pm 1.5}$ & $\mathbf{36.4 \pm 1.6}$ \\
CaLLiPer-Llama & $\underline{34.5 \pm 1.0}$ & $\underline{34.6 \pm 1.0}$ & $\underline{33.8 \pm 0.9}$ & $\mathbf{38.8 \pm 1.0}$ & $\underline{35.8 \pm 0.85}$ & $\underline{35.5 \pm 1.0}$ \\
\bottomrule
\end{tabular}
\end{table}

The results of the LUC task are presented in Table \ref{tab:luc_res}. We can observe that the embeddings produced by CaLLiPer (both variants) surpass other competitive methods across all evaluation metrics in both downstream models (i.e., Linear and MLP). This indicates that our method captures the spatial distributions of urban functions implied by POIs more effectively, and this information significantly aids in land use classification. Moreover, our method outperforms other baselines by a large margin in the linear probing setting, further proving that the pre-trained representations are highly robust and thus do not require complex downstream models or extensive supervised training. 

As for the performance of the other compared methods, the probabilistic topic modelling models, i.e., TF-IDF and Doc2Vec, show limited improvement when using a more complex non-linear model, indicating that these models may not capture sufficient spatial or contextual nuances from the POI data, limiting their ability to benefit from non-linear transformations. In contrast, deep representation learning models (Place2Vec, Doc2Vec, SPPE) that learn POI embeddings first and then aggregate, achieved significantly better performance with the MLP model compared to the linear probing setting, suggesting that they perform more effectively when combined with non-linear models. HGI, which uses a pre-defined delineation of the urban space as LSOA, performed worse than SPPE (the model it is based on) with MLP as the downstream model, indicating that relying on rigid spatial divisions limits its ability to generalise across varying urban contexts and capture finer spatial patterns. As for Space2Vec, despite its location encoding capability, it did not perform well in this task. We argue that this is mainly because the encoder-decoder architecture falls short in learning the geographic prior as effectively as our proposed model. 

\subsubsection{Socioeconomic status distribution mapping}

Table \ref{tab:sdm_res} shows the performance on the SDM task. CaLLiPer models consistently achieved the best performance across all metrics. This indicates that the distributions estimated using the CaLLiPer representations have the smallest distance and the greatest similarity to real distributions. It is worth noting that CaLLiPer-Llama outperformed CaLLiPer-SenTrans on the SDM task but performed worse on the LUC task (see Table \ref{tab:luc_res}). This suggests that using a larger text encoder may not always provide additional benefits. Moreover, the discrepancy between these two model variants is not significant. Therefore, in real-world applications where efficiency is prioritised, more lightweight text encoders are preferable.

The performance of the other compared models is similar to what was observed in the LUC task (presented in Section \ref{sec:luc_res}), except that HGI performed significantly better on the SDM task. This improvement is likely because the target SES distribution is measured on LSOAs, which is the spatial delineation used to train the HGI model. This in turn, echoed the first issue mentioned in Section 1, where existing methods are limited by the predefined spatial delineation of urban space.
In summary, the results in both the LUC and the SDM task clearly demonstrate the effectiveness of the CaLLiPer model in learning useful representations of the urban environment, benefiting various downstream urban-related tasks.

\begin{table}[]
\caption{Performance comparison on the SDM task. The best and second-best performance are marked in \textbf{bold} and \underline{underlined}, respectively. For better readability, all metrics are scaled by a factor of $10^{-2}$.}
\begin{tabular}{l|ccc|ccc}
\toprule
\multirow{2}{*}{Model} & \multicolumn{3}{c|}{Linear} & \multicolumn{3}{c}{MLP} \\
 & L1 ↓ & Chebyshev ↓ & KL ↓ & L1 ↓ & Chebyshev ↓ & KL ↓ \\
\midrule
Random & 30.31 ± 0.03 & 9.25 ± 0.01 & 7.73 ± 0.01 & 31.40 ± 0.22 & 9.55 ± 0.11 & 8.21 ± 0.14 \\
TF-IDF & 24.79 ± 0.04 & 7.43 ± 0.01 & 5.36 ± 0.01 & 24.36 ± 0.15 & 7.28 ± 0.05 & 5.20 ± 0.04 \\
LDA & 26.14 ± 0.01 & 7.84 ± 0.00 & 5.87 ± 0.00 & 25.85 ± 0.14 & 7.77 ± 1.12 & 5.80 ± 0.72 \\
Place2Vec & 23.47 ± 0.09 & 6.94 ± 0.02 & 4.81 ± 0.02 & 22.81 ± 0.06 & 6.81 ± 0.01 & 4.61 ± 0.02 \\
Doc2Vec & 24.01 ± 0.07 & 7.15 ± 0.02 & 4.99 ± 0.02 & 23.10 ± 0.19 & 6.89 ± 0.06 & 4.75 ± 0.08 \\
SPPE & 24.32 ± 0.16 & 7.24 ± 0.06 & 5.11 ± 0.06 & 23.63 ± 0.19 & 7.04 ± 0.06 & 4.91 ± 0.07 \\
HGI & 23.28 ± 0.08 & 6.93 ± 0.02 & 4.79 ± 0.03 & 22.73 ± 0.05 & 6.80 ± 0.02 & 4.60 ± 0.02 \\
Space2Vec & 25.13 ± 0.15 & 7.56 ± 0.04 & 5.65 ± 0.06 & 23.55 ± 0.20 & 7.12 ± 0.09 & 5.00 ± 0.08 \\
\midrule
CaLLiPer-SenTrans & \underline{22.21 ± 0.12} & \underline{6.71 ± 0.36} & \underline{4.45 ± 0.03} & \underline{21.65 ± 0.12} & \underline{6.59 ± 0.02} & \underline{4.32 ± 0.06} \\
CaLLiPer-Llama & \textbf{21.89 ± 0.05} & \textbf{6.64 ± 0.02} & \textbf{4.30 ± 0.03} & \textbf{21.23 ± 0.22} & \textbf{6.43 ± 0.02} & \textbf{4.09 ± 0.05} \\
\bottomrule
\end{tabular}
\label{tab:sdm_res}
\end{table}

\subsection{Qualitative results and analyses}

We conduct qualitative analysis of the learned representations in the following sections. Unless stated otherwise, the representations produced by CaLLiPer-SenTrans are used throughout.

\subsubsection{Clustering of the learned representations}

\begin{figure}
    \centering
    \includegraphics[width=0.96\linewidth]{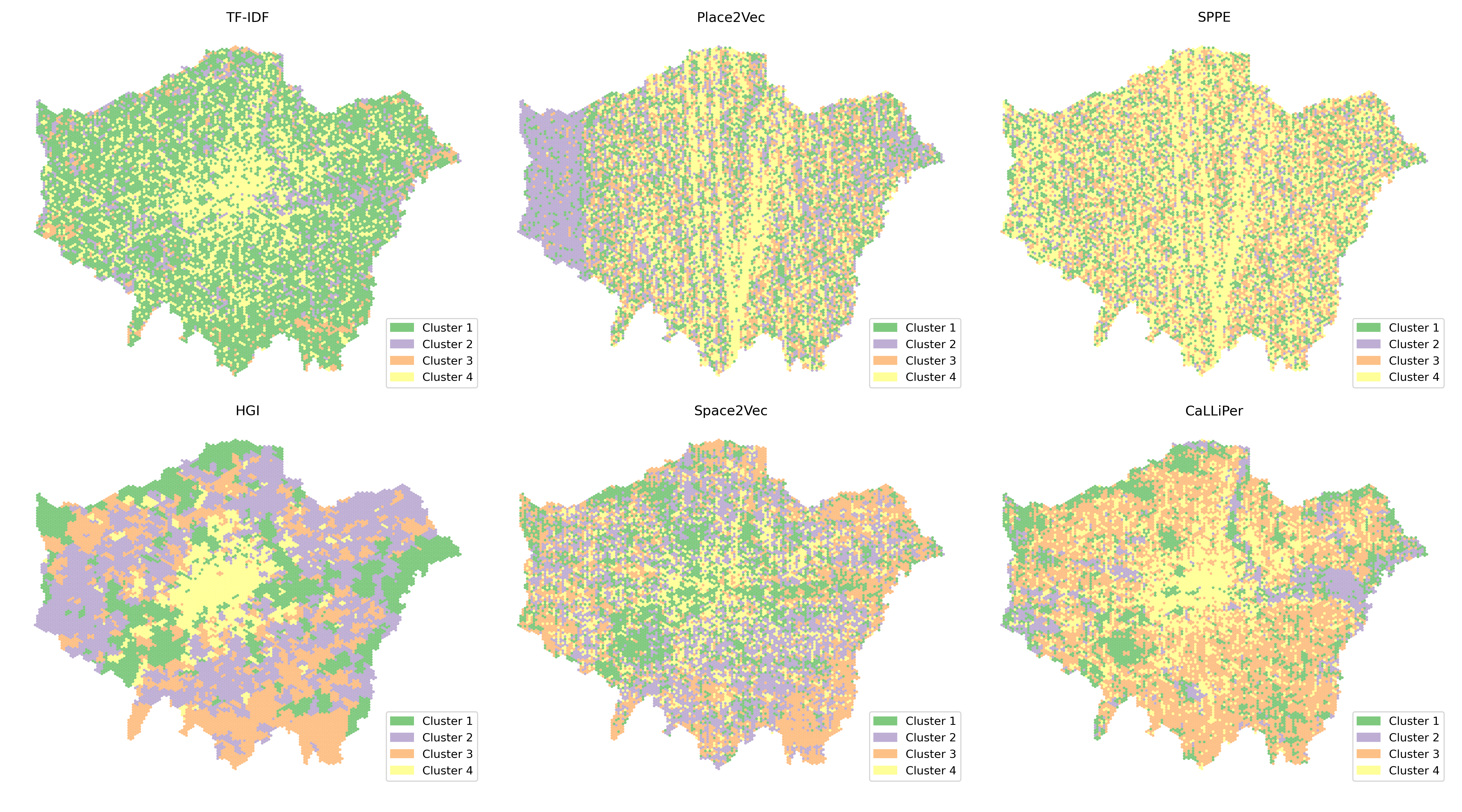}
    \caption{Clustering results of the representations for the hexagonal grid of London.}
    \label{fig:hex_cluster}
\end{figure}

To further investigate the effectiveness of the compared methods, we conducted clustering analysis and presented the results in Figure \ref{fig:hex_cluster}. Specifically, we generated the representations for each hexagonal area in London hexagonal grid\footnote{https://data.london.gov.uk/dataset/green-infrastructure-focus-map}  using six of the compared methods. We adopted basic K-Means clustering for this analysis, following the practice of \citep{niu2021delineating}, although other clustering methods like hierarchical clustering could also be applied \citep{mai2020iclr}. From the visualisation, we can observe that there are no clear spatial distribution patterns for Place2Vec and SPPE. While some spatial patterns are visible in TF-IDF and Space2Vec results, they appear more dispersed and random compared to CaLLiPer’s. The clusters produced from HGI representations are more spatially contiguous and less noisy but have coarser spatial resolution because of the larger spatial units (i.e., LSOAs) originally used for training the model. 

\begin{figure}
    \centering
    \includegraphics[width=0.9\linewidth]{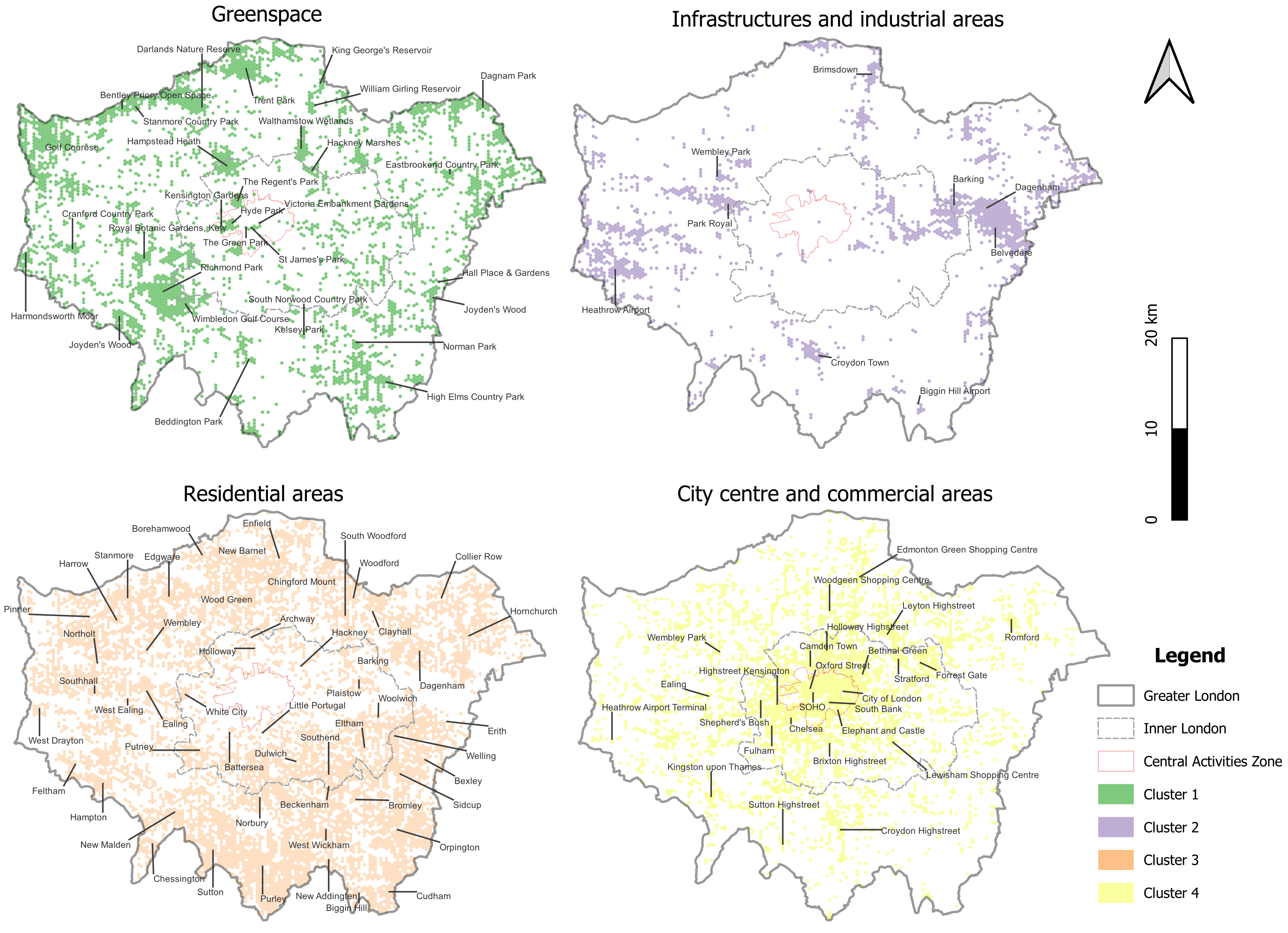}
    \caption{Clustering results of CaLLiPer-SenTrans representations, which are partially labelled by the names of greenspace, landmarks, local neighbourhoods/towns, high streets, etc.}
    \label{fig:hex_cluster_interpret}
\end{figure}

To help interpret the results, we further visualised the clustering results of CaLLiPer representations and labelled the prominent local areas in Figure \ref{fig:hex_cluster_interpret}. As shown in the figure, the 4 clusters can be interpreted as ``Greenspace'', “Infrastructures and Industrial Areas”, “Residential Areas” and “City Centre and Commercial Areas”, respectively. Each cluster overlaps well with the actual urban functional zones. Such a clear classification of urban functional areas was not seen in any of the other methods, further verifying the superior capacity of our method in accurately capturing and differentiating key urban spatial structures with fine spaital resolutions and high semantic fidelity. 

\subsubsection{Zoom into local areas}

\begin{figure}
    \centering
    \includegraphics[width=0.98\linewidth]{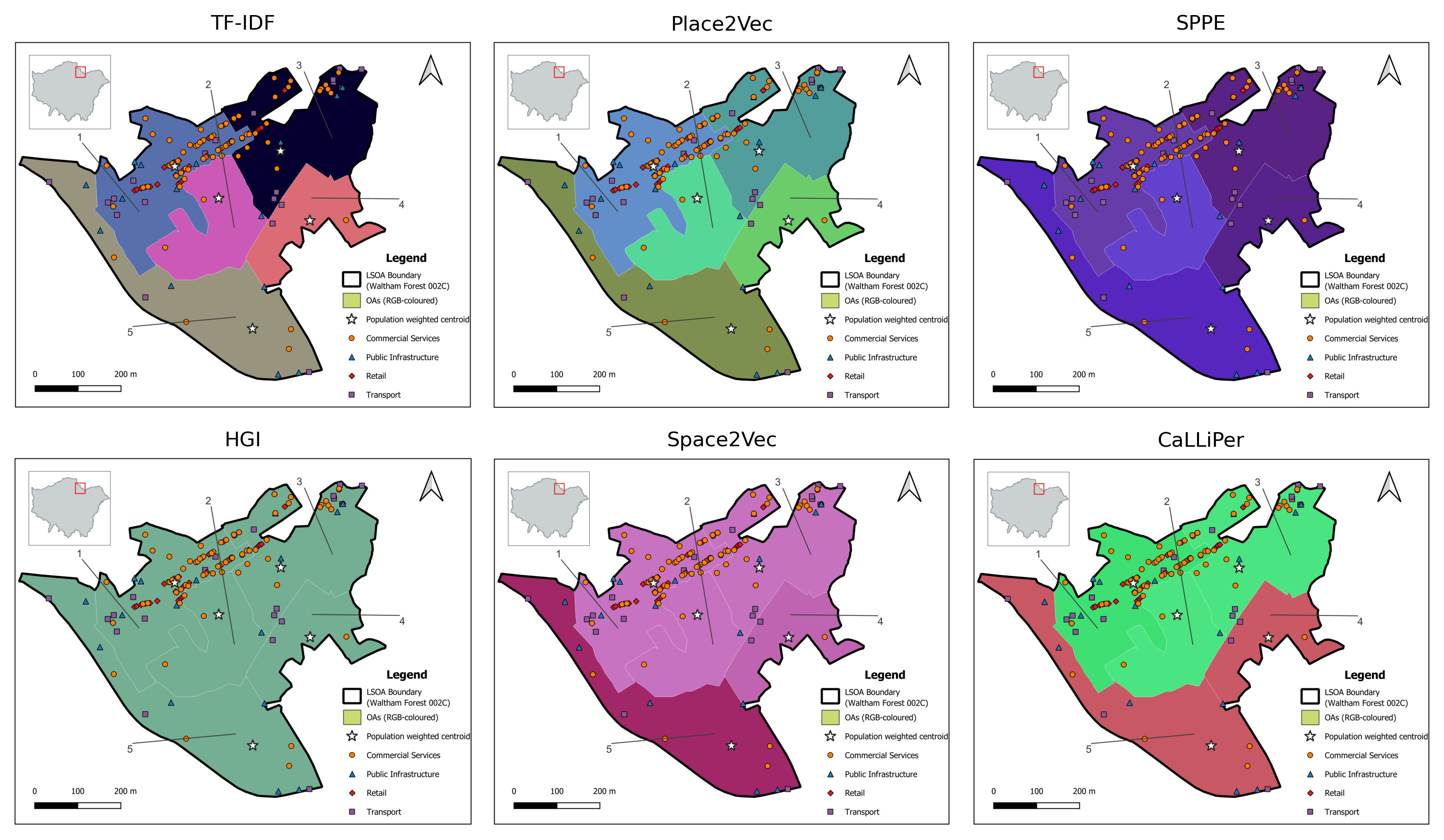}
    \caption{Latent representation visualisation in a local area. The RGB colours of OAs come from the 3-dimensional vectors derived from original representations via t-SNE.}
    \label{fig:local_area}
\end{figure}

In this subsection, we visualise a low-dimensional projection of the latent representations of London’s output areas (OAs) , which are subdivisions of LSOAs. For better comparison, we focus on a specific local area—LSOA Waltham Forest 002C—as a case study. Specifically, we projected the 128-dimensional representation vectors into 3-dimensional vectors using t-SNE and used these for RGB-colouring of the OAs in Figure \ref{fig:local_area}. From the figure, we can observe that OA 1 (labelled as 1 in the figure; the other OAs similarly labelled) contains many POIs, especially commercial services, while the other OAs contain significantly fewer POIs. Geographically, OA 2 and 3 are closer to OA 1, while OA 4 and 5 are farther away.

Looking at the colouring of OAs, we can see that HGI represented each sub-area of the LSOA with a single representation due to the predetermined spatial division of urban space, failing to provide representations with a finer resolution than the level it was originally trained on. For models strictly confined by the spatial division of OAs, i.e., TF-IDF, Place2Vec, and SPPE, they coloured OA 1 and 2 quite differently even though they are adjacent (with many POIs located near the border), indicating that their representations in the vector space are dissimilar. This does not fully reflect the real-world situation. In contrast, location encoding-based models, such as Space2Vec and CaLLiPer, effectively account for the spatial proximity of OA 1, 2, and 3. Our CaLLiPer model further distinguishes OA 4 from OA 1, 2, and 3 because OA 4 is farther away, offering a more nuanced understanding of the area.

\subsection{Computational efficiency}
\label{sec:6.3}

We present the computation time for all neural network-based models in Table \ref{tab:compute_time} to compare their computational efficiency. To ensure a fair comparison, the batch size is set to 1024 for all methods except for SPPE, which uses a batch size of 32768. This is because using a batch size of 1024 for SPPE would be nearly intractable due to the significantly longer training time required (refer to Appendix \ref{sec:appendix_imp_sppe_hgi} for more details). 

It is evident that simpler models, like the word2vec-based Place2Vec, are much faster to train. SPPE has the highest per-epoch training time, primarily because it involves constructing a training corpus through random walks on a weighted graph generated by Delaunay Triangulation of tens of thousands of POIs, leading to a much larger training set compared to the other methods. HGI has the second highest per-epoch training time due to the sophisticated negative sampling strategy it uses in its contrastive learning process. Additionally, HGI requires many training epochs, which results in an extremely long total training time. Our model demonstrates relatively high training efficiency, with each epoch taking about 20 seconds and the entire training process completed within half an hour. This makes it both efficient and accurate, making it a more practical choice for real-world applications. 

\begin{table}[]
\caption{Comparison of computation time.}
\label{tab:compute_time}
\begin{tabular}{lrr}
\toprule
Model & Training time per epoch (seconds) & Total training time (minutes) \\
\midrule
Place2Vec & 7.8 & 2.0 \\
SPPE & 263.9 & 439.8 \\
HGI & 66.9 & 2230.7 \\
Space2Vec & 16.8 & 18.0 \\
CaLLiPer & 19.5 & 20.5 \\
\bottomrule
\end{tabular}
\end{table}

\section{Conclusions and discussion}
\label{sec:conclustion_discussion}

In this paper, we presented a novel approach, CaLLiPer, for learning urban space representations based on POIs using a contrastive location-language pre-training objective. The proposed method offers a unique perspective by directly embedding the continuous geographical space into a higher-dimensional vector space to capture the spatial distribution of the semantics of the urban environment. We combined location encoding with contrastive learning to learn the geographic prior of the entire study area, effectively capturing the spatial variations in POI semantic distributions. We used pre-trained text encoders from the NLP domain to fully capture the semantic meaning of POIs, intricately aligning the location embedding space with the language embedding space. Comprehensive experiments on two downstream tasks demonstrated the effectiveness of our method, confirming its effectiveness and potential for a wide range of urban studies. In the following subsections, we delve into a more in-depth discussion to interpret the findings and explore their broad implications.

\subsection{Capturing spatial variations}

The extensive experiments demonstrated the superior performance of our model. But why is our model so successful? We argue that it is because our model addresses the core challenge in learning representations for computational urban studies. Many past studies have emphasised that the core of learning urban space representations lies in capturing the spatial variations of POI distributions \citep{zhai2019beyond,huang2023hgi}. Our model aligns location embeddings with their corresponding semantic representations derived from POIs, which closely matches with this objective, and thus enabling the improved performance. 

In comparison, existing models might be inherently suboptimal for learning urban space representations. For deep representation learning methods that first learn POI category embedding and then aggregate them to specific areas, this indirect modelling of the distribution can result in the loss of valuable information (i.e., spatial variations), especially during the aggregation process. On the other hand, modelling urban spaces by treating each discrete area as a node in a graph structure incurs problems such as ecological fallacy (due to the scales of the predetermined discrete areas) and a lack of generalisability, as shown by the experimental results of HGI. 

\subsection{Implications for geospatial foundation models}

Our model tackles the fundamental problem of geospatial data representation learning. It combines location encoding with semantic encoding of POIs through contrastive learning, fusing spatial (coordinates) and platial (place attributes) information by answering “what is where.” \citep{goodchild2020platial}. As a result, our model enables spatially explicit modelling \citep{janowicz2020geoai} of the semantics of urban environments.

The resulting representations of geographical spaces have the potential to be applied to nearly all geospatial tasks, especially those relying on accurate characterisations of the urban environments. For example, apart from the tasks that have been conducted in our experiment, some other direct applications of our learned representations include tasks such as individual’s next location prediction \citep{hong2023context} or forecasting the visitation flow of people \citep{feng2017poi2vec}, given that human mobility is largely associated with the semantic meaning of urban spaces. And as such, we argue that our proposed model represents a step towards geospatial foundation models \citep{xie2023geo,mai2024opportunities}.  

Moving forward, our model could be applied to other cities and regions by adding POI data in other cities in the training. Future research efforts could also be made in further improving the location-semantic encoding and incorporating other modalities like imagery data. 


\appendix


\section{More implementation details}

\subsection{Prompt templates for CaLLiPer}
\label{sec:appendix_prompt_template}

As described in Section \ref{sec:method_text_encoder}, we used prompt template to enhance the modelling of POI semantics. Specifically, we used the Group and Class labelling and constructed a rather simple prompt template – “A place of {Class label}, a type of {Group label}”. To account for special descriptions of some POIs, for example, some Class-level descriptions of Retail POIs are the actual goods, in this case, we use “A place that \textbf{sells} \{Class label\}, a type of \{Group label\}”. Similar modification applies to “Manufacturing and Production” POIs, for which we use “A place that \textbf{produces} \{Class label\}, a type of \{Group label\}”. 

\subsection{Implementation details of SPPE and HGI}
\label{sec:appendix_imp_sppe_hgi}

Following the formulation of SPPE \citep{huang2022sppe}, we constructed the POI network through Delauney Triangulation. The resulting POI network comprised 259,254 nodes and 777,737 edges, which is a very large network. We used the default settings for conducting the biased random walk, setting the number of walks per node as 5, which resulted in 1,296,270 random walks. The large amount of training samples forced us to use a large batch size (32,768), but even so, it still resulted in a rather long training time (refer to Section \ref{sec:6.3}). As SPPE does not have an early stopping mechanism, we trained the model for 100 epochs and saved a checkpoint every 10 epochs, and we evaluated each checkpoint on both LUC and SDM task and found the best performing one was at the 90th epoch, whose performance was reported in Section \ref{sec:luc_res}. 

In our experiment of the HGI model \citep{huang2023hgi}, we took the pre-trained POI category embeddings from SPPE (the 90-epoch checkpoint) as the corresponding POI category embeddings in HGI. Then we follow the default setting of the original paper and train the model for 2000 epochs, which incur an extraordinary amount of training time. 

\subsection{Implementation details of Space2Vec}
\label{sec:appendix_imp_space2vec}

The original space2vec paper \citep{mai2020iclr} implements two types of decoders: one is location decoder (used for location modelling) and the other is spatial context decoder (used for spatial context modelling). To better support our task, we used the GlobalPositionEncoderDecoder in the experiment. We constructed the negative samples of each POI by randomly selecting 10 classes that are different from the centre POI’s. To facilitate fair comparison between Space2Vec and our proposed CaLLiPer model, we kept the hyperparameter settings of their location encoders consistent. 

\section{More experimental results}
\subsection{Other location encoding methods}

As mentioned in Section \ref{sec:lit_loc_encoding}, various location encoding methods exist. To assess the impact of using different location encoding methods, we conducted additional experiments in which only the location encoder of CaLLiPer-SenTrans was modified, while the other components remained unchanged. The performance results on both the LUC and SDM tasks are presented in Table \ref{tab:appendix_loc_enc_res}.

\begin{table*}[]
\caption{Performance comparison of model variants using different location encoders. The best performances are highlighted in \textbf{bold}.}
\label{tab:appendix_loc_enc_res}
\begin{tabular}{l|ccc|ccc}
\toprule
\multirow{2}{*}{Location encoder} & \multicolumn{3}{c|}{LUC} & \multicolumn{3}{c}{SDM} \\
 & Precision ↑ & Recall ↑ & F1 score ↑ & L1 ↓ & Chebyshev ↓ & KL ↓ \\
 \midrule
Grid & 38.4 ± 1.5 & \textbf{36.8 $\pm$ 1.5} & \textbf{36.4 $\pm$ 1.6} & 21.65 ± 0.12 & 6.59 ± 0.02 & 4.32 ± 0.06 \\
Theory & \textbf{38.5 $\pm$ 1.0} & 36.6 ± 0.8 & 36.2 ± 0.5 & 21.15 ± 0.12 & 6.33 ± 0.04 & 4.12 ± 0.08 \\
SphereC & 18.1 ± 1.1 & 19.4 ± 0.6 & 16.8 ± 0.8 & 25.2 ± 0.08 & 7.51 ± 0.03 & 5.57 ± 0.04 \\
Spherical Harmonics & 37.7 ± 1.6 & 32.9 ± 1.5 & 32.4 ± 1.5 & \textbf{18.31 $\pm$ 0.35} & \textbf{5.56 $\pm$ 0.09} & \textbf{3.26 $\pm$ 0.09} \\
\bottomrule
\end{tabular}
\end{table*}

We can observe that the Grid \citep{mai2020iclr} encoder achieved similar performance to the Theory \citep{mai2020iclr} encoder, which conforms to what we anticipated. As for SphereC \citep{mai2023sphere2vec} and Spherical Harmonics (SH) \citep{russwurm2024sh}, both are designed for location encoding on spherical surfaces. SH achieved the best performance on the SDM task but did not perform well on the LUC task. In contrast, SphereC showed the lowest performance on both tasks.

\subsection{Comparing with SatCLIP}
\label{sec:appendix_satclip}

As mentioned in Section \ref{sec:2.4}, SatCLIP \citep{klemmer2023satclip} used satellite images for contrastive image-location pre-training. Here, we used a pre-trained checkpoint of SatCLIP\footnote{https://huggingface.co/microsoft/SatCLIP-ViT16-L40} in the downstream tasks. This rough comparison serves the purpose of comparing the effect of the two modalities (text V.S. satellite images). 

From Table \ref{tab:compare_satclip}, the representations generated from CaLLiPer achieved better performance, roughly corroborated our hypothesis that supervision from POIs is more suitable for learning urban space representations possibly due to their unique advantages of providing more detailed semantic information with finer resolutions. 

\begin{table}[]
\caption{Comparison with SatCLIP.}
\label{tab:compare_satclip}
\begin{tabular}{l|ccc|ccc}
\toprule
\multirow{2}{*}{Model} & \multicolumn{3}{c|}{LUC} & \multicolumn{3}{c}{SDM} \\
 & Precision ↑ & Recall ↑ & F1 score ↑ & L1 ↓ & Chebyshev ↓ & KL ↓ \\
 \midrule
CaLLiPer-SenTrans & 38.4 ± 1.5 & 36.8 ± 1.5 & 36.4 ± 1.6 & 21.65 ± 0.12 & 6.59 ± 0.02 & 4.32 ± 0.06 \\
SatCLIP & 12.2 ± 2.5 & 16.8 ± 2.0 & 12.2 ± 1.6 & 28.89 ± 0.09 & 8.84 ± 0.03 & 7.07 ± 0.01 \\
\bottomrule
\end{tabular}
\end{table}




\printcredits

\bibliographystyle{cas-model2-names}

\bibliography{xlw_refs}



\end{document}